\documentclass{article}

     \usepackage[preprint]{neurips_2019}

\usepackage[utf8]{inputenc} 
\usepackage[T1]{fontenc}    
\usepackage{hyperref}       
\usepackage{url}            
\usepackage{amsmath}
\usepackage{amsfonts}       
\usepackage{nicefrac}       
\usepackage{microtype}      
\usepackage{graphicx}
\graphicspath{ {./figures/} }
\usepackage{caption}
\usepackage{subcaption}
\usepackage{booktabs} 
\usepackage{tikz}
\usetikzlibrary{arrows,shapes,trees,backgrounds}

\title{Deep density ratio estimation for change point detection}

\author{%
  Haidar Khan \\
  Department of Computer Science\\
  Rensselaer Polytechnic Institute\\
  Troy, NY 12180 \\  
  \texttt{khanh2@rpi.edu} \\
  \And
  Lara Marcuse \\
  The Mount Sinai Epilepsy Center \\
  Mount Sinai Hospital \\
  New York, NY 10029 \\
  \texttt{lara.marcuse@mssm.edu} \\
  \And
  B\"{u}lent Yener \\
  Department of Computer Science \\
  Rensselaer Polytechnic Institute \\
  Troy, NY 12180 \\
  \texttt{yener@rpi.edu} \\
}

\begin{document}

\maketitle

\begin{abstract}
  In this work, we propose new objective functions to train deep neural network based density ratio estimators and apply it to a change point detection problem. Existing methods use linear combinations of kernels to approximate the density ratio function by solving a convex constrained minimization problem. Approximating the density ratio function using a deep neural network requires defining a suitable objective function to optimize. We formulate and compare objective functions that can be minimized using gradient descent and show that the network can effectively learn to approximate the density ratio function. Using our deep density ratio estimation objective function results in better performance on a seizure detection task than other (kernel and neural network based) density ratio estimation methods and other window-based change point detection algorithms. We also show that the method can still support other neural network architectures, such as convolutional networks. 
\end{abstract}

\section{Introduction}

The need for change point detection arises in many contexts; such as computer network intrusion detection~\citep{Yamaishj2002}, meteorological events~\citep{Reeves2007},  and speech recognition~\citep{Rybach2009}. Change point detection is challenging due to a number of factors including noise in the underlying data, nonlinearity in the time-series, and the (generally) unknown generating distributions.

The setting we study here involves systems that generate data we can observe as a time-series. The exact parameters and model of the systems involved are generally unknown but we are aware that the system transitions from one state to another. The change point detection problem is determining when (in time) the transition between states occurs.

\begin{figure}[h]
	\centering
	\includegraphics[width=1\columnwidth]{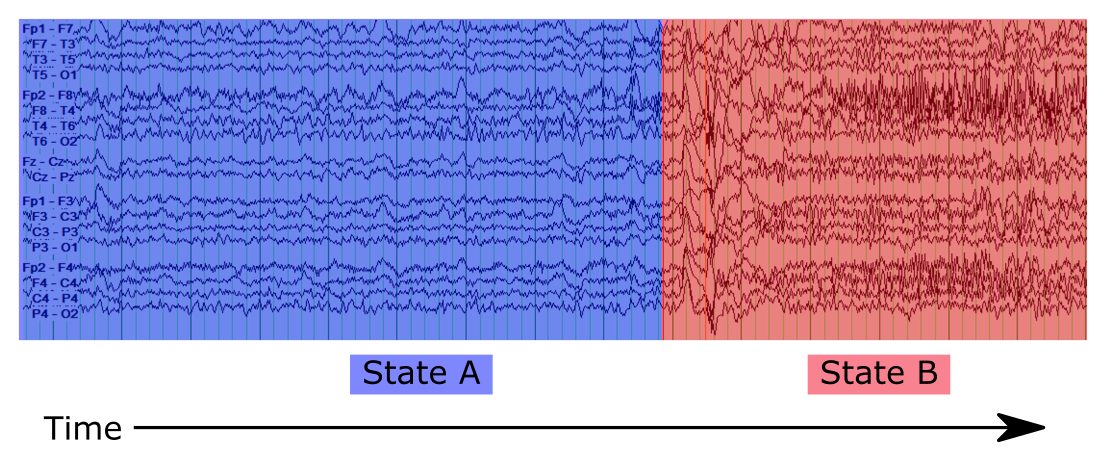}
	\caption{Scalp EEG recording of a patient experiencing a seizure. The time series captures the transition of the patient from the non-seizure state (State A) to the seizure state (State B). The change point detection problem is determining the time at which the transition occurs.}
	\label{sig_states}
\end{figure}

\subsection{Problem Definition}

The variant of the change point detection (CPD) problem considered here is stated as follows. Given a set of time series $\{X_i\}_{i=1}^n$ where each $X_i$ is a set of observations, or feature vectors, at discrete time steps $X_i=\{x_1,x_2,...x_{T_i}\}$ with $x_t \in \mathbb{R}^d$, we assume each $X_i$ is generated by a system that undergoes a transition from state $A$ to state $B$ as shown in Figure~\ref{sig_states}. The CPD problem is determining the time at which the transition occurs in the time series $X_i$ represented by a time $\tau_i$. Suppose $P_A(x)$ and $P_B(x)$ are the unknown probability distributions of the observations over the space $\mathcal{X}=\mathbb{R}^d$ and that $P_A(x) \ne P_B(x)$. A change point is defined as a time $\tau$ such that:

\[\{x_1,x_2,...,x_\tau\} \sim  P_A (x)\]
\[\{x_{\tau+1},x_{t+2},...,x_{T_i} \} \sim P_B(x)\]

If the distributions $P_A(x)$ or $P_B(x)$ are known or easy to estimate, the CPD problem can be reduced to a simple hypothesis testing framework. Because this is not often the case, an array of methods have been developed for the CPD problem.

The rest of this paper is organized as follows. In Section~\ref{cpd_prior}, we review change point detection methods. In Section~\ref{cpd_dre} density ratio estimation and its application to change point detection is discussed in more detail. Section~\ref{cpd_ddre} introduces density ratio estimation using deep neural networks and two novel objective functions for training networks for density ratio estimation. The empirical results are presented in Section~\ref{cpd_results}. We discuss the results and conclude in Section~\ref{cpd_disc} and Section~\ref{cpd_conc} respectively.

\section{Related Work}
\label{cpd_prior}
Early work on CPD posed the problem as comparing the probability distributions of time series over successive intervals to detect changes~\citep{Basseville1994}. A hypothesis testing framework was typically adopted with the null hypothesis of “no change” and the alternative hypothesis of “a change occurred”. Examples of this approach include the cumulative sum (CUSUM) algorithm~\citep{Basseville1994}, the generalized likelihood ratio~\citep{Gustafsson1997}, and likelihood detectors~\citep{Kuncheva2013}. The CUSUM algorithm calculates a cumulative sum of deviations from a baseline model over a time series and detects a change point when the sum passes a threshold. The generalized likelihood ratio test assumes a linear state space model and uses the likelihood ratio instead of the likelihood for hypothesis testing. Likelihood detectors estimate the probability distributions with Gaussian mixtures. These methods work well in single or low dimensional spaces, but are unable to detect change points in high dimensional time series.


Parametric methods relying on model assumptions such as auto regressive (AR) models ~\citep{Yamaishj2002} and subspace models~\citep{Ide2007} have also been proposed. The change finder ~\citep{Yamaishj2002} measures deviations from an AR model with linearity assumptions to detect outliers and change points. Subspace methods~\citep{Ide2007} detect change points using distances between the subspaces of past and present windows estimated with principal component analysis. These methods demonstrate success for change point detection but are ineffective when model assumptions are violated.

Kernel methods for estimating the probability densities over consecutive intervals are an example of non-parametric models for change point detection. The kernel Fischer discriminant ratio has been proposed to detect change points~\citep{Mika1999,Harchaoui2009}. Kernel Mean Matching~\citep{Gretton2009} matches the moments of the distributions in a reproducing kernel Hilbert space with a kernel, such as the Gaussian kernel, and measures the distances between the distributions over separate intervals. These methods rely on the choice of kernel function instead of a pre-defined model but are still inaccurate when the feature space is high-dimensional~\citep{Vapnik1998}.

Attempts to circumvent density estimation in high dimensional spaces with non-parametric methods have focused on using kernels to estimate the ratio of the distributions instead of the probability distributions themselves. This line of research, called density ratio estimation, is motivated by the idea that since knowing the densities of two intervals implies knowing the density ratio but that knowing the ratio does not imply knowing the individual densities, estimating the ratio is an easier problem~\citep{Kawahara2012}. The KL divergence based importance estimation procedure (KLIEP) ~\citep{Suigyama2007,Kawahara2012} is one example of this approach. KLIEP estimates the density ratio of two consecutive samples using Gaussian kernels and detects change points when the density ratio exceeds a threshold. A similar approach is proposed with the unconstrained least squares importance fitting (uLSIF) and relative uLSIF~\citep{Liu2013}.

Virtual classifier based methods use classification algorithms to detect change points in time series~\citep{Desobry2005,Hido2008}. 
This method has been extended to feature selection~\citep{Yamada2013} using the Hilbert Schmidt Independence Criterion for feature ranking. These methods show promise for dealing with change point detection in high-dimensional spaces. Drawbacks include the prohibitive computational cost for training and testing a classifier for each candidate change point as well as a strong dependence on the feature space of the time series.

In this work, we improve density ratio estimation by using deep neural networks as the density ratio function approximator. This approach leverages the layered representation learning capabilities of deep networks to learn better approximations to the density ratio. The improved density ratio estimation results in better performance in change point detection.

\section{Density Ratio Estimation}
\label{cpd_dre}
The ratio of two probability densities arises in many contexts such as Monte Carlo importance sampling and covariate shift detection. The ratio is useful for change point detection when it is taken between the distribution of the feature vectors $x$ over the state space $\mathcal{X} = \mathbb{R}^d$ conditioned on the state $B$ and the state $A$: \[ \beta = \frac{P(x|B)}{P(x|A)} \] 

For a time series, the density ratio $\beta_t$ of a sample $x_t$ measures the likelihood the sample comes from the distribution of state $B$ vs. the distribution of state $A$. When viewed as a weight corresponding to each sample, a high value for $\beta_t$ means the sample is likely to be from $B$ and a value close to zero indicates the sample is from $A$. In the ideal case, if a change from state $A$ to state $B$ occurs in a time series $X_i$ then the $\beta$ values act like a step function over time where the step occurs at the change point.

The challenge with this approach is that $\beta$ is not easy to compute. Directly estimating the probability densities $P(x|A)$ and $P(x|B)$ is difficult because of the general difficulty of density estimation in high dimensions. Since we only require the density ratio, estimating the conditional probability densities is actually over-solving the problem. We can avoid this by attempting to estimate the density ratio directly with a parameterized function and optimizing the parameters to match the true density ratio. Suppose we estimate the density ratio with a function $f(x;\theta)$ with parameters $\theta$: $ f(x;\theta)\approx \frac{P(x|B)}{P(x|A)} $

The parameters $\theta$ can be optimized by noting that $P(x|B)=P(x|A)f(x;\theta)$ and minimizing the KL divergence between $P(x|B)$ and $P(x|A)f(x;\theta)$, as in the KLIEP method~\citep{Suigyama2007}:

\[ 
\min_\theta{\mbox{KL}(P(x|B),P(x|A)f(x;\theta)}
\]
\[ \min_\theta{ \sum_{x\in\mathcal{X}}{P(x|B)\log{\frac{P(x|B)}{P(x|A)}}}-\sum_{x\in\mathcal{X}}{P(x|B)\log{f(x;\theta)}} }\]

Since the first term does not depend on $\theta$, this can be rewritten as a minimization over the second term:

\[ \min_\theta{ -\sum_{x\in\mathcal{X}}{P(x|B)\log{f(x;\theta)}} }\]

Equality constraints are used to ensure $P(x|A)f(x;\theta)$ represents a valid probability distribution:

\begin{equation*}
	\sum_{x\in\mathcal{X}}{P(x|A)f(x;\theta)} = 1
\end{equation*}

The function used to approximate the density ratio can be any parameterized function, such as a weighted linear combination of kernel functions: 
\[f(x;\theta)=\sum_{i=1}^b{\theta_iK(x_i,x)}\]
In the case of functions which are linear in the parameters $\theta$ the optimization problem is convex. For a neural network approximator $f(x;\theta)=\sigma(\theta_1 \sigma(\theta_2 x))$ (with nonlinear activation function $\sigma(\cdot)$) the optimization is not straightforward.

Estimating the density ratio for change point detection on a dataset of time series requires using the empirical estimates for the summations expressed above as well as splitting each time series into reference and evaluation segments. Each time series is split by assuming the system starts generating the time series in state $A$. Thus, the first $n_{r}$ samples of each time series are labeled as being generated in state $A$. The remaining $n_{e}$ samples of each time series are unlabeled, belonging either to state $A$ or state $B$. This setting is referred to as semi-supervised or positive/unlabeled data in which only the reference samples are labeled (i.e. $n_r$ is given). The task is to determine the label of the samples in the evaluation segment as well as previously unseen samples.

After each time series is split into reference and evaluation segments, the optimization problem can be expressed as:

\begin{equation}
	\begin{split}
		\min_\theta{ -\frac{1}{n_{e}} \sum_{i=1}^{n_{e}}{\log{f(x;\theta)}} } \\
		f(x_j;\theta) > 0 \forall j=1...n_{r} \\
		\frac{1}{n_{r}} \sum_{j=1}^{n_{r}}{f(x_j;\theta)} = 1  \\
	\end{split}
	\label{kliep_eq}
\end{equation}

Density ratio estimation provides a number of advantages over other methods when applied to change point detection. Each sample in the evaluation segment is given a weight that represents how different the sample is from the samples in the reference distribution without testing multiple possible change points. This is similar to methods for detecting covariate shift between training and test datasets, where each sample in the training dataset is given a weight signifying how similar or different it is to the test dataset distribution. 

A drawback to the density ratio estimation methods presented here is that, as referenced earlier, it is not easy to optimize the parameters of the density ratio approximator when the objective is a non-linear function of the parameters $\theta$. This is the case when using a deep neural network to approximate the density ratio. The motivation to use a deep neural network for the density ratio approximator is that the layers in the neural network allow learning representations of the data at the same time as optimizing the objective. In the next section, we describe modifications to the above formulation that allows approximation of the density ratio with a deep neural network.

\section{Deep Density Ratio Estimation}
\label{cpd_ddre}
One of the attractive qualities of the virtual classifiers method for change point detection is that it allows the optimization of both the feature space and the change point through the use of deep neural networks. However, this comes at a large computational cost because of the need to test multiple candidate change points. Density ratio estimation on the other hand, can directly weight evaluation samples based on their similarity to a reference distribution. The drawback to this is that the feature space flexibility is limited. We attempt to combine the strengths of both of these methods using deep neural networks to approximate the density ratio.

The main challenge to this approach is selecting an appropriate objective function to train the network to approximate the density ratio. The objective given in Equation~\ref{kliep_eq} cannot be directly optimized by a network because of the constraints. Training a good neural network based approximator to the density ratio requires an objective function that satisfies the constraints and can be optimized using minibatch stochastic gradient descent.

\subsection{LSIF Objective}

This approach has been explored in~\citep{Nam2015} using a modified version of the least squares importance fitting (\textbf{LSIF})~\citep{Liu2013} objective function to train a convolutional network to approximate the density ratio. This formulation of the objective function is given by:

\begin{equation}
	\frac{1}{n_{e}} \sum_{j=1}^{n_{e}}{f(x_j^{e};\theta)^2} - \frac{2}{n_{r}} \sum_{i=1}^{n_{r}}{f(x_i^{r};\theta)}
	\label{cnn_lsif}
\end{equation}

In the formulation given in ~\citep{Nam2015}, the network is trained with pairs of reference and evaluation samples. However, this results in minibatch sizes too small to train on large datasets. We show in our experiments that using larger minibatch sizes with this objective function results in poor performance for change point detection. 

We propose two new formulations of the density ratio estimation objective that allow approximating the density ratio with a deep neural network (DNN).

\subsection{DSKL Objective}

Our first formulation uses the KL-divergence in both directions, also referred to as the JS-divergence, to avoid the equality constraints. This is motivated by considering the behavior of the density ratio estimator when the equality constraint in Equation~\ref{kliep_eq} is dropped:

\[ \min_\theta{ -\frac{1}{n_{e}} \sum_{i=1}^{n_{e}}{\log{f(x;\theta)}} }  \]
\[ f(x_j;\theta) > 0 \forall j=1...n_{r} \]

Since the objective only depends on the evaluation samples, $f(x;\theta)$ is free to assign arbitrarily large values to the density ratio for the evaluation samples with no regard to the effect on the reference samples. To avoid this, we consider minimizing the KL-divergence between $P(x|B)$ and $P(x|A)f(x;\theta)$ and the KL-divergence between $P(x|A)$ and $\frac{P(x|B)}{f(x;\theta)}$. This yields:

\[ \min_\theta{ \mbox{KL}(P(x|B),P(x|A)f(x;\theta))+ \mbox{KL}(P(x|A),\frac{P(x|B)}{f(x;\theta)}) } \]

\[ \min_\theta{ \sum_{x\in\mathcal{X}}{P(x|B)\log{\frac{P(x|B)}{P(x|A)f(x;\theta)}}}
+\sum_{x\in\mathcal{X}}{P(x|A)\log{\frac{P(x|A)f(x;\theta)}{P(x|B)}} }  }
\]

Dropping the terms irrelevant to the optimization because they do not depend on $\theta$:

\[ 
\min_\theta{
	-\sum_{x\in\mathcal{X}}{P(x|B)\log{f(x;\theta)}}
	+\sum_{x\in\mathcal{X}}{P(x|A)\log{f(x;\theta)}}
}
\]

Replacing the summations with the empirical estimates and assuming a reference and evaluation split yields the \textbf{DSKL} objective:

\begin{equation}
	-\frac{1}{n_{e}}\sum_{i=1}^{n_{e}}{\log{f(x_i;\theta)}}
	+\frac{1}{n_{r}}\sum_{j=1}^{n_{r}}{\log{f(x_j;\theta)}}	
	\label{dskl_obj}
\end{equation}

The function $f(x;\theta)$ is an approximator to the density ratio, and therefore it does not represent a valid probability distribution. The density ratio is nonzero but unbounded from above. In this minimization problem, we are concerned with both samples where $f(x;\theta)<1$ (where the probability distribution in the denominator is larger than the distribution in the numerator) and where $f(x;\theta)>1$. 

The objective function in Equation~\ref{dskl_obj} can be minimized using full-batch gradient descent to optimize the parameters of the neural network. When using minibatch gradient descent or stochastic gradient descent the minibatch size must be optimized as a hyperparameter. Approximately equal numbers of reference and evaluation samples should be included in each batch.

\subsection{BARR Objective}

The second formulation uses a different approach to handle the constraints in Equation~\ref{kliep_eq}. A Lagrange multiplier is used to incorporate the equality constraint into the objective function as a barrier term. The original objective function can then be rewritten as the \textbf{BARR} objective:

\begin{equation}
	-\frac{1}{n_{e}}\sum_{i=1}^{n_{e}}{\log{f(x_i;\theta)}}
	+ \lambda \left| \frac{1}{n_{r}}\sum_{j=1}^{n_{r}}{f(x_j;\theta)} - 1  \right| 	
	\label{barr_obj}
\end{equation}

The Lagrange multiplier $\lambda$ plays the familiar role of a regularizer in this formulation. Choosing the value for $\lambda$ can be guided by the same heuristics used to choose regularization parameters for the model. Optimizing this objective by minibatch or stochastic gradient descent does not guarantee the equality constraint will be satisfied, especially when $\lambda$ is small. For our experiments, the value of $\lambda$ was fixed at 10.

We propose using a neural network to approximate the density ratio and tune the parameters of the network using each of the objective functions. This will allow learning a feature space at the same time as estimating when the system transitions from one state to the next.

\section{Results}
\label{cpd_results}
In our experiments, we compared the change point detection performance of the proposed method, called the \textbf{deep density ratio estimator (DDRE)} trained using the objective functions given in Equations~\ref{cnn_lsif},~\ref{dskl_obj}, and ~\ref{barr_obj}, to two other popular density ratio estimation methods, KLIEP~\citep{Suigyama2007} and rULSIF~\citep{Liu2013} as well as other change point detection algorithms. We give a brief description of each change point detection (CPD) method tested in the appendix.

The metric used to compare methods is average detection lag, also known as average run length. For a given set of true change points $\{t\}_{i=1}^N$ and a set of corresponding predicted change points $\{\tau\}_{i=1}^N$, the average detection lag is $ADL = \frac{1}{N}\sum_{i=1}^{N}{|t_i-\tau_i|}$. 

\subsection{Simulations}
Monte Carlo simulations were carried out to estimate average detection lag for three methods, KLIEP, rUSLIF, and DDRE. A gaussian distribution was used to model the underlying time series and the change point was modeled by a change in the parameters of the distribution. This experiment required two generating distributions; $P(x|A)$ and $P(x|B)$, the pre-change and post-change distributions respectively. We varied the dimensionality of the feature space ($D$) from 10-100. Each experiment was repeated for 20 iterations. 

We used a multivariate gaussian distribution $\mathcal{N}(\mu_1, \Sigma_1)$ with $\mu_1 \in \mathbb{R}^D$ and $\Sigma_1 \in \mathbb{R}^{D \times D}$ randomly selected from a uniform distribution ($\Sigma_1$ was constrained to be a symmetric positive definite matrix) for $P(x|A)$. $P(x|B)$ was generated by perturbing the parameters of $P(x|A)$ by random amounts.

%

\subsubsection{Simulation results}
The deep density ratio estimator (DDRE) used in our experiments consisted of 5 fully connected layers with ReLU activations ~\citep{Nair2010} and 500 units in each layer except the last layer. The final output activation of this network is also a ReLU because the density ratio takes values in the range $[0,\infty)$. Dropout ~\citep{Srivastava2014} and L-2 regularization of $0.01$ was applied at every layer. The network was trained using the DSKL loss function (Equation~\ref{dskl_obj}). Mini-batch stochastic gradient descent was used with a batch size of 200 and an initial learning rate of 1e-3. In each experiment the network was trained until the validation loss stopped decreasing.

The average detection lags (ADL) for each method are shown in Figure~\ref{sim_res}, a lower ADL is better. Each of these figures shows the estimated ADL of KLIEP, rULSIF, and DDRE. We observed that the performance gap between DDRE and the other methods increased with the dimension of the feature space, indicating DDRE functions better than other methods in high dimensional spaces. This is due to the deep neural networks automatic feature extraction capabilities.

\subsection{EEG Dataset}

In the epileptic brain, seizures are characterized by abnormal firing of large networks of neurons. Since seizures can occur as random times, detecting the onset of a seizure is useful for patients, caregivers, clinicians, and even neuro-stimulation devices. The electrical activity of the brain can be monitored externally using a device containing electrodes placed on the scalp, called an electroencephalogram (EEG). The most widely used configuration of the electrodes on the scalp is known as the 10-20 system. 


The EEG dataset contains 121 recordings with 22 channels sampled at 256 Hz each containing a single lead seizure. The EEG data is part of a publicly available dataset recorded at the Mount Sinai Hospital \footnote{\url{http://www.dsrc.rpi.edu/?page=databank}}. The data is provided in the European Data Format~\citep{Kemp2003}. Statistics of the dataset are summarized in the appendix.

\begin{figure}[h]
	\centering
	\begin{subfigure}{0.48\linewidth}
		\centering
		\includegraphics[width=\linewidth]{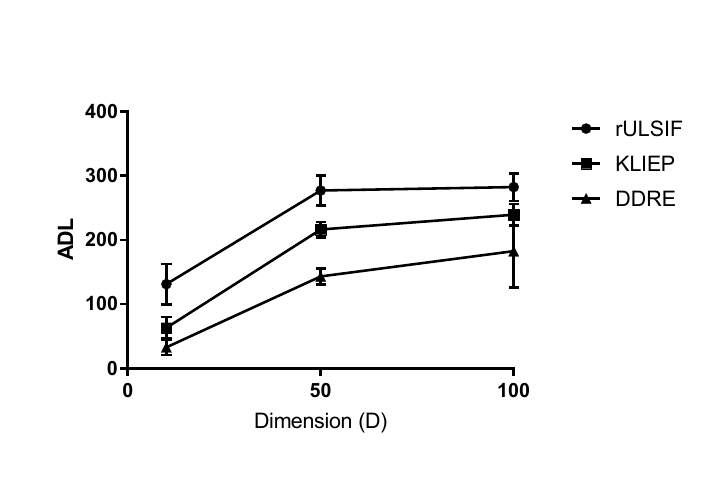}
		\caption{Shown are the simulation results for each of the three density ratio estimation methods. The deep density ratio estimator (DDRE) outperforms the other methods (lower ADL is better). Error bars show one standard deviation.}
		\label{sim_res}
	\end{subfigure}
	\begin{subfigure}{0.48\linewidth}
		\centering
		\begin{tikzpicture}[scale=0.95]
		\foreach \y in {1,1.4,1.8,2.2,2.6}
		\draw[fill=blue, very thick] (1,\y ) rectangle (2.5,\y+.3);
		\foreach \y in {1,1.4,1.8,2.2,2.6}
		\draw[left color=blue, right color=red, very thick] (2.5,\y ) rectangle (5,\y+.3);
		
		\draw (0.7,0.9) -- (0.5,0.9) -- (0.5,2.91) -- (0.7,2.91);
		\draw (0.4,1.905) node[left] {Train} -- (0.5, 1.905);
		\draw (1.2,2.9) node[above right] {ref};
		\draw (3,2.9) node[above right] {eval};
		
		\draw[shade, very thick] (1,.5 ) rectangle (5,.5+.3);
		\draw[shade, very thick] (1,.1 ) rectangle (5,.1+.3);
		\draw (0.7,0.09) -- (0.5,0.09) -- (0.5,0.86) -- (0.7, 0.86);
		\draw (0.4,.5) node[left] {Test} -- (0.5, .5);
		\end{tikzpicture}
		\caption{Diagram showing how the database of time series is split. Each time series contains a single change point. The reference segments represent sections of the time series known to be in state A. The evaluation segments represent sections of the time series with unknown state labels.}
		\label{setup}
	\end{subfigure}
	\caption{ }
\end{figure}

The data in each of our experiments consisted of a dataset of $N=121$ time series, each of which contains an actual change from a state $A$ (nonseizure) to a state $B$ (seizure). For each experiment the dataset of time series is first split into a test set (20\%) and a training set (80\%). Since all the recordings contain a minimum of 60 minutes of nonseizure activity before the seizure, the first 20\% of samples in each time series in the training set are used to create the reference segments. The remaining samples in each time series in the training set are allocated as the evaluation segment. The model is then trained on the reference and evaluation samples using cross-validation to select parameters. The resulting model is used to estimate the density ratios of the samples in the test set. This setup is shown in Figure~\ref{setup}.

When using the density ratio estimation approach, the model outputs the estimated density ratio for each sample in each time series in the test set. These are transformed into a change point prediction for each time series by fitting a sigmoid function such as the logistic function to the ratio values with parameters $k$ and $x_0$: $ y = -\frac{1}{1 + e^{-k(t-x_0)}} $  
The fitted value of the parameter $x_0$ is used as the predicted change point. This method was adopted as the thresholding function for all the density ratio estimation methods. This is not necessary for window based change point detection methods as the algorithm directly returns the most likely change point.


\begin{figure}[h]
	\centering
	\begin{subfigure}[t]{0.48\linewidth}
		\centering
		\includegraphics[width=0.85\linewidth]{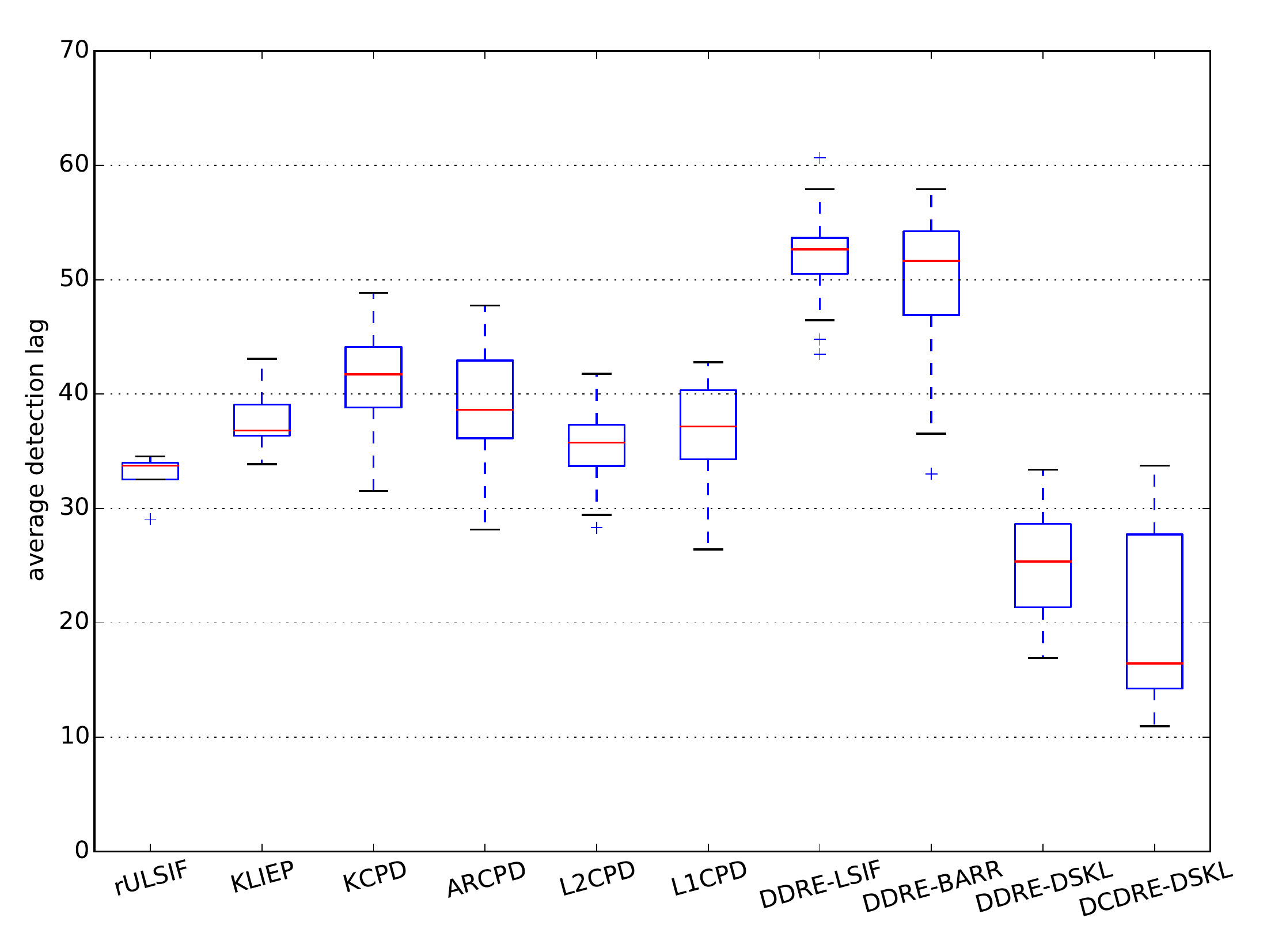}
		\caption{Average detection lag (ADL) scores (in seconds) for each method. 30 bootstrap resampling runs were used to construct box plots of the ADL. The red horizontal line shows the median ADL for each method.}
		\label{seizure_res}
	\end{subfigure}
	~
	\begin{subfigure}[t]{0.48\linewidth}
		\centering
		\includegraphics[width=0.75\linewidth]{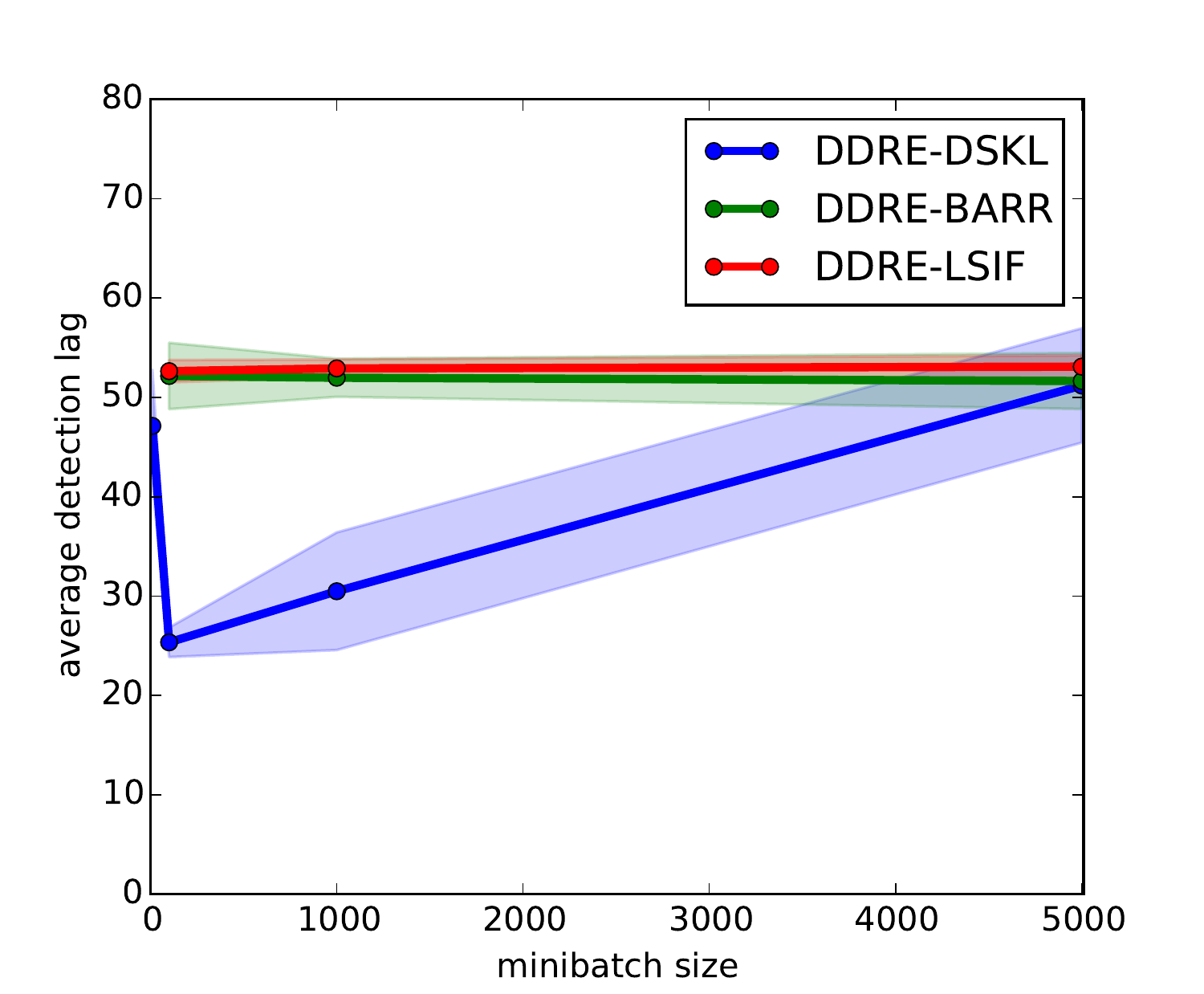}
		\caption{The effects of minibatch size on average detection lag for the three objective functions LSIF, BARR, and DSKL. The LSIF and BARR objectives were unstable for small minibatch sizes (10)}
		\label{minibatch}	
	\end{subfigure}
	\caption{Empricial results on EEG dataset}
\end{figure}


\section{Discussion}
\label{cpd_disc}
We compare the performance of density ratio based change point detection methods in detecting seizure onset on this dataset. All methods except DCDRE were trained on 396 signal features extracted from the EEG. The features consisted of three energy measures (average energy, Teager-Kaiser energy, and line-length) computed on six frequency bands for each channel as described in~\citet{Cook2013b}. The convolutional neural network (CNN) was trained directly on the wavelet transform of the EEG signal.

The average detection lag for each method over 30 bootstrap resampling runs is shown in Figure~\ref{seizure_res}. The kernel based density ratio estimation methods (rULSIF and KLIEP) and window based change point detection methods (*CPD) achieve 31-42 median ADL (lower is better). Using kernel functions to approximate the density ratio did not show much improvement over window based change point detection algorithms.

DDRE-LSIF and DDRE-BARR, two deep density ratio estimation methods, perform much worse at 51-52 ADL. DDRE-DSKL outperforms the other density ratio estimation methods and window based change point detection method with a lower ADL of 25.
Figure~\ref{minibatch} shows the effect of minibatch size on ADL. Unlike the other two density ratio estimation objectives LSIF and BARR, the DSKL objective treats reference and evaluation samples equally allowing the network to update the density ratio on the evaluation samples using the (given) information about the reference set.

In all cases, the deep convolutional density ratio estimator (DCDRE) outperformed the other methods (16 ADL) but exhibited high variance across bootstrapping runs. This improvement is due to the ability to incorporate the feature extraction capabilities of the convolutional filters into density ratio estimation. The high variance of the deep density ratio estimators compared to the kernel based density ratio estimators is caused by the tendency of deep neural network optimization to converge to local optima. 


\section{Conclusion}
\label{cpd_conc}
In this work, we develop novel objective functions for density ratio estimation using deep neural networks and compare their performance on a seizure detection task. Our proposed method uses a modified objective function suited to minibatch gradient descent training. We compare our method to two state of the art kernel-based density ratio estimation methods as well as change point detection methods. The results show a clear advantage to deep neural network based density ratio estimation methods especially when convolutional networks are used to learn feature representations.

%

\medskip
\small
\bibliographystyle{plainnat}
\bibliography{C:/Users/Haidar/Documents/library}

\clearpage
\section*{Appendix}

\subsection*{Density Ratio Based CPD}

These methods perform change point detection by first learning an approximation to the density ratio function. During inference, the learned function is applied to the input signal and the change point is determined using a thresholding function.

\paragraph{KLIEP}
The Kullback-Leibler Importance Estimation Procedure is a density ratio estimation method that uses kernel functions, typically gaussian kernels, to approximate the density ratio. Specifically, the density ratio is approximated by a weighted combination of kernels:
\[
f(x;\theta) = \sum_{l=1}^{b}{\theta_l \phi(x, x_l)}
\]
The parameters $\theta$ are learned by solving the convex optimization problem given in Equation~\ref{kliep_eq} which minimizes the KL-divergence between the reference distribution and the approximation ~\cite{Suigyama2007}.

\paragraph{rULSIF}
Similar to KLIEP, Relative Unconstrained Least Squares Importance Fitting uses kernels to approximate the density ratio, however a different optimization problem is solved to learn the parameters. The objective function is derived using the squared loss and is written as:

\[
\frac{1}{2 n_r} \sum_{x\in\mathcal{X}}{f(x;\theta)^2} - \frac{1}{n_e}\sum_{x\in\mathcal{X}}{f(x;\theta)} 
\]

For a complete description, see~\citet{Liu2013}.\footnote{We used the Matlab code for KLIEP and rULSIF provided by the authors at \url{http://www.ms.k.u-tokyo.ac.jp/software.html}}

\paragraph{DDRE}
Deep Density Ration Estimation uses a deep feedforward network as a function approximator to the density ratio. We compare three objective functions for training the deep feedforward network:

\begin{itemize}
	\item DDRE-LSIF: trained using the LSIF objective (Equation~\ref{cnn_lsif})~\citep{Nam2015}.
	\item DDRE-DSKL: trained using the DSKL objective (Equation~\ref{dskl_obj}).
	\item DDRE-BARR: trained using the BARR objective (Equation~\ref{barr_obj}).
\end{itemize}

For each of these objective functions, we use a five layer rectified linear unit (ReLU) network~\cite{Nair2010}. The first four layers are 500 units wide each. The output layer consists of one unit with a ReLU activation function as the density ratio function is nonnegative. L2 regularization with $\lambda=0.01$ was applied to each layer as well as Dropout~\cite{Srivastava2014} with keep probability 0.5. The network was trained using the ADAM~\cite{Kingma2015} optimizer and minibatch gradient descent. Training was continued until the objective stopped decreasing, typically 50-100 epochs. This network structure was fixed in all of our experiments to ensure fair comparison between the objective functions. Our implementation was written using Tensorflow~\citep{Abadi2016a}. 

\paragraph{DCDRE}
We also experimented with a convolutional neural network using the DSKL objective function. We borrowed the architecture of this network from the CNN trained on EEG data in~\citet{Khan2017} except for the last layer. The network consists of six convolutional and max-pooling layers followed by three fully connected layers ending in a single unit. The convolutional and pooling structure of the network is shown in Table~\ref{arch_table}. The network contains ReLU activation functions at every layer as well as L2 regularization ($\lambda=0.01$) and Dropout ($p=0.25$). The ADAM optimizer and minibatch gradient descent ($bS=100$) were used to train the network.

\begin{table}[h]
	\caption{Convolutional network architecture for EEG data~\cite{Khan2017}.}
	\label{arch_table}
	\centering
	\begin{tabular}{cccc}
		Type & Units & Size & Stride\\
		\midrule
		
		Conv2D & 60 & 3x3 & 1\\
		\midrule
		Pool2D & & 2x2 & 1\\
		\midrule
		
		Conv2D & 50 & 3x3 & 1\\
		\midrule
		
		Conv2D & 40 & 3x3 & 1\\
		\midrule
		Pool2D & & 2x2 & 1\\
		
		\midrule
		Conv2D & 20 & 3x3 & 1\\
		\midrule
		Pool2D & & 2x2 & 1\\
		
		\midrule
		Conv2D & 10 & 2x2 & 1\\
		
		\midrule
		Conv2D & 5 & 2x2 & 1\\
		
		\midrule
		Fully Conn. & 250 & & \\
		
		\midrule
		Fully Conn. & 100 & & \\
		
		\midrule
		Output & 1 & & \\
		
		\bottomrule
	\end{tabular}
\end{table}

\subsection*{Window Based CPD}
These methods perform change point detection using two sliding windows along the signal. A discrepancy measure is computed at each time point using a cost function. Peaks in the discrepancy measure are considered change points. Each of the following methods defines a different cost function for computing the discrepancy. Following is a short description of each, for details see~\citep{Truong2018}. \footnote{We used the python code provided by the authors in our experiments \url{https://github.com/deepcharles/ruptures}}

\paragraph{KCPD}
Kernel Change Point Detection algorithms use a kernel function that maps the signal onto a reproducing kernel Hilbert space to measure the cost. A typical choice of kernel function is the Gaussian kernel.

\paragraph{ARCPD}
Auto Regressive Change Point Detection uses deviations or residuals from a linear auto regressive model to model the cost between adjacent windows.

\paragraph{L2CPD}
L2 Change Point Detection uses the Euclidean distance to measure the cost.

\paragraph{L1CPD}
L1 Change Point Detection uses the L1 distance as the cost function.

\begin{table}[h]
	\caption{Mount Sinai Hospital EEG dataset statistics.}
	\label{dataset_table}
	\centering
	\begin{tabular}{lc}
		\toprule
		\# Recordings & 121 \\ 
		Average Length & 1.55h \\
		\# Channels & 22 \\
		\# Features & 396 \\
		Window Length & 5s \\
		\# Windows & 143803 \\			
		\bottomrule
	\end{tabular}
\end{table}

\end{document}